\documentclass[10pt,twocolumn,letterpaper]{article}

\usepackage[pagenumbers]{cvpr} 
\usepackage{graphicx}
\usepackage{amsmath}
\usepackage{amssymb}
\usepackage{booktabs}
\usepackage{multirow}
\usepackage{amsfonts}
\usepackage{algorithm}
\usepackage{algorithmic}
\usepackage{bm}
\usepackage[dvipsnames]{xcolor}
\usepackage{colortbl}  
\usepackage{xcolor}

\usepackage[accsupp]{axessibility}

\usepackage[pagebackref=true,breaklinks=true,letterpaper=true,colorlinks,citecolor=ForestGreen, bookmarks=false]{hyperref}

\usepackage[capitalize]{cleveref}
\crefname{section}{Sec.}{Secs.}
\Crefname{section}{Section}{Sections}
\Crefname{table}{Table}{Tables}
\crefname{table}{Tab.}{Tabs.}

\newcolumntype{*}{>{\global\let\currentrowstyle\relax}}
\newcolumntype{^}{>{\currentrowstyle}}

\definecolor{dt}{gray}{0.7}  %

\definecolor{myred}{RGB}{223,211,112}
\definecolor{mygreen}{RGB}{156,199,227}

\def\confName{CVPR}
\def\confYear{2024}
\newcommand{\authorskip}{\hspace{5mm}}

\begin{document}

\title{SimDA: Simple Diffusion Adapter for Efficient Video Generation}

\author{Zhen Xing\textsuperscript{1} \authorskip Qi Dai\textsuperscript{2} \authorskip Han Hu\textsuperscript{2}  \authorskip Zuxuan Wu\textsuperscript{1} \authorskip Yu-Gang Jiang\textsuperscript{1} \\[0.5mm]
{
\textsuperscript{1}  Shanghai Key Lab of Intell. Info. Processing, School of CS, Fudan University}\\ 
{\textsuperscript{2} Microsoft Research Asia}
}

\twocolumn[{
\maketitle
\vspace{-1.4em}
\renewcommand\twocolumn[1][]{#1}
\begin{center}
    \centering
    \includegraphics[width=0.94\textwidth]{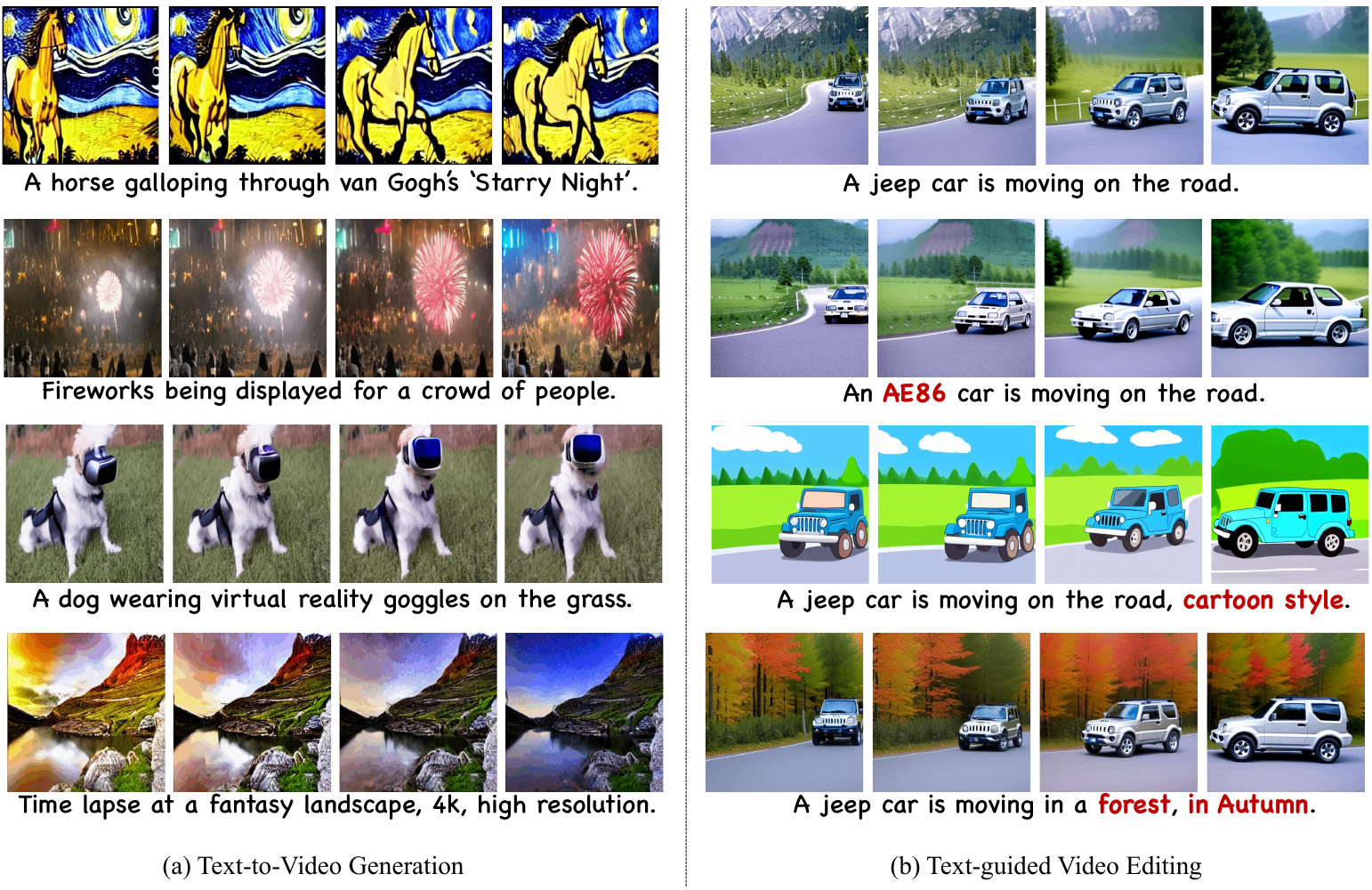}
    \vspace{-0.0cm}
    \captionof{figure}{
    \textit{\textbf{Examples of our SimDA results:}} (a) The results of open-wild Text-to-Video Generation. (b) Text-guided video Editing results using one text-video pair tuning.
    }
    \label{fig:fig1}
    \vspace{-0.0cm}
\end{center}
}]

\begin{abstract}
The recent wave of AI-generated content has witnessed the great development and success of Text-to-Image (T2I) technologies.
By contrast, Text-to-Video (T2V) still falls short of expectations though attracting increasing interests.
Existing works either train from scratch or adapt large T2I model to videos, both of which are computation and resource expensive.
In this work, we propose a Simple Diffusion Adapter (SimDA) that fine-tunes only 24M out of 1.1B parameters of a strong T2I model, adapting it to video generation in a parameter-efficient way.
In particular, we turn the T2I model for T2V by designing light-weight spatial and temporal adapters for transfer learning. Besides, we change the original spatial attention to the proposed Latent-Shift Attention (LSA) for temporal consistency. With similar model architecture, we further train a video super-resolution model to generate high-definition ($1024 \times 1024$) videos. In addition to T2V generation in the wild, SimDA could also be utilized in one-shot video editing with only 2 minutes tuning. Doing so, our method could minimize the training effort with extremely few tunable parameters for model adaptation.
More examples can be found at our website \url{https://chenhsing.github.io/SimDA/}. 
\end{abstract}

\section{Introduction}

Image generation stands on top of the recent AIGC wave.
It not only has a significant impact in the academic community but also achieves tremendous success in various applications, such as computer graphics, art and culture, medical imaging, \emph{etc}. The approaches in this area mainly include methods based on generative adversarial networks (GANs)~\cite{gan,aliasgan,stylegan,stylegan2,stylegan3}, auto-regressive transformers~\cite{taming,zeroshot,scaling}, and the latest diffusion models~\cite{diffusionbeatgan,ddpm,improvedddpm,cascaded,glide,dalle,stablediffusion,imagen, song2020score,feng2023ernie,liu2022compositional,xue2023raphael}.
Among them, diffusion models are the most popular owing to the strong controllability, simple stability, and amazing realism. 
However, video generation research lags behind due to challenges like the scarcity of publicly available datasets, difficulty in modeling temporal information, and high training costs, hindering the progress in this area.

There have been several research endeavors dedicated to exploring video synthesis~\cite{babaeizadeh2017stochastic,brooks2022generating,castrejon2019improved,denton2018stochastic,franceschi2020stochastic,ge2022long,gupta2022rv,gupta2018imagine, hong2022cogvideo, kahembwe2020lower,lee2018stochastic, li2018video, luc2020transformation, mittal2017sync, marwah2017attentive, pan2017create, saito2020train, skorokhodov2022stylegan, tian2021good, villegas2017decomposing, vondrick2016generating,weissenborn2019scaling,wu2021godiva, yan2021videogpt, yu2022generating}. In addition, some studies have employed popular diffusion models for video generation~\cite{harvey2022flexible,vdm,hoppe2022diffusion,voleti2022masked,yang2022diffusion, videofusion}. However, most of them involve training models from scratch, which can be time-consuming due to the complex video data. Early attempts were also constrained by GPU memory or hardware limitations.

More recently, a small number of T2V (Text-to-Video) approaches have emerged, aiming to fine-tune well-established T2I (Text-to-Image) models~\cite{ho2022imagenvideo, videoLDM, latentshift,videofactory, videofusion, magicvideo, preserve}. 
They have incorporated temporal modeling modules (\emph{e.g.} Imagen video~\cite{ho2022imagenvideo}, Video LDM~\cite{videoLDM}) into
T2I models, which effectively accelerate the model convergence.
However, it should be noted that training such models is still a challenging task due to the massive number of parameters (4B or even 16B) involved in the network architecture. 

In the NLP field, state-of-the-art results of various tasks are generally achieved by adaptation from large pretrained models (\emph{i.e.}, BERT~\cite{bert}, LLMs~\cite{dai2019transformer,radford2019language, raffel2020exploring, zhang2022opt}).
However, with the advent of increasingly larger and more powerful foundation models (\emph{e.g.}, GPT-4 with 100T parameters), conducting full fine-tuning of the entire models has become prohibitively expensive and infeasible in terms of training cost and GPU storage. 
To address the issue, numerous methods based on efficient fine-tuning have emerged rapidly in NLP~\cite{houlsby2019parameter, hu2021lora, li2021prefix, lester2021power} and computer vision~\cite{st-adapter, yang2023aim, multimodaladapter, chen2022adaptformer}.

In this work, we propose a parameter-efficient video diffusion model, namely Simple Diffusion Adapter (SimDA), that fine-tunes the large T2I (\emph{i.e.} Stable Diffusion~\cite{stablediffusion}) model for improved video generation. We only add $0.02\%$ parameters compared to the T2I model. During training, we freeze the original T2I model, and only tune the newly added modules. 
We further propose a Latent-Shift Attention (LSA) to replace the original spatial attention, which significantly improves the temporal modeling capability and retains consistency without adding new parameters.
To this end, our model demands less than 8GB GPU memory for training with a resolution of $16\times256\times256$, while the inference time speeds up by $~39\times$ compared to the auto-regressive method CogVideo~\cite{hong2022cogvideo}. 
Besides, we turn an image super resolution framework into the video counterpart with similar architecture, which allows generating high-definition videos of $1024\times1024$. 
Our model can also be extended to the recently popular diffusion-based video editing~\cite{tuneavideo}, achieving significant $3\times$ faster training while retaining comparable results, as evidenced by the editing examples presented in Fig~\ref{fig:fig1} (b).
In conclusion, the contributions of this work can be summarized as follows:
\begin{itemize}
\item We explore the simple adaptation from image diffusion to video diffusion, exhibiting that tuning extremely few parameters can achieve surprisingly good results.
\item With the helpful light-weight adapters and the proposed latent-shift attention, our method can effectively model the temporal relations with negligible cost.
\item Our diffusion adapter could be extended to text-guided video super-resolution and video editing, significantly facilitating the model training.
\item SimDA significantly alleviates the training cost and speeds up the inference time, while remaining competitive results compared to other methods.

\end{itemize}

\section{Related Work}

\paragraph{Text-to-Video Generation}

Similar to the advancements in Text-to-Image (T2I) generation, early approaches for Text-to-Video (T2V) generation~\cite{li2018video, mittal2017sync, pan2017create} were based on Generative Adversarial Networks (GANs) and primarily applied to domain-specific videos such as simple human actions~\cite{ucf101} or clouds moving~\cite{time-lapse}. Due to the inherent challenges of video data modeling and the requirements for large-scale high-quality text-video datasets, the development of open-wild T2V generation is limited. However, learning a prior from T2I generation can effectively alleviate this problem.

For instance, NÜWA~\cite{wu2022nuwa} formulates a unified representation space for images and videos, enabling multitask learning for both T2I and T2V generation. CogVideo~\cite{hong2022cogvideo} incorporates temporal attention layers into the pretrained and frozen CogView2~\cite{ding2022cogview2} model to capture motion dynamics. Make-A-Video \cite{singer2022make} proposes fine-tuning a pretrained DALLE2~\cite{dalle} model solely on video data to learn motion patterns, enabling T2V generation without explicitly training on text-video pairs. Video Diffusion Models~\cite{vdm} and Imagen Video~\cite{ho2022imagenvideo} perform joint text-image and text-video training, treating images as independent frames and disabling temporal layers in the U-Net~\cite{unet} architecture. Phenaki~\cite{villegas2022phenaki} also conducts joint training for T2I and T2V generation using the Transformer model, considering an image as a frozen video. Besides, Video LDM~\cite{videoLDM}, Latent-Shift~\cite{latentshift}, VideoFactory~\cite{videofactory}, MagicVideo~\cite{magicvideo} and our methods utilize the popular open-sourced T2I Stable Diffusion~\cite{stablediffusion} model. While the progress in video generation is impressive, the parameters of video generation can be highly large. As shown in Table~\ref{tab:model_size}, Make-A-Video~\cite{singer2022make} require six models and 9.7B parameters and Imagen Video~\cite{ho2022imagenvideo} utilizes eight models with 16.3B parameters, which limits the training efficiency of T2V models.

\paragraph{Text guided Video Editing}
In the realm of content generation, an alternative avenue is the manipulation of existing images~\cite{brooks2023instructpix2pix, hertz2022prompt2prompt, meng2021sdedit, tumanyan2023plug} and videos~\cite{tuneavideo,bar2022text2live, gen1, liu2023video, qi2023fatezero, shin2023edit,zhang2023towards, yang2023rerender} using textual input as a means of control, rather than relying solely on unbridled text-based generation. SDEdit~\cite{meng2021sdedit} introduces noise to images and then reconstructs them for the purpose of editing. Prompt-to-prompt~\cite{hertz2022prompt2prompt} and Plug-and-Play~\cite{tumanyan2023plug} modify the cross-attention map by altering the textual description, thus influencing the editing process. When it comes to video editing, Tune-A-Video~\cite{tuneavideo} fine-tunes the T2I (Text-to-Image) model on a single video, enabling the generation of new videos with similar motion patterns. Video-P2P~\cite{liu2023video} and FateZero~\cite{qi2023fatezero} extend the concept of Prompt-to-prompt editing to the realm of videos. Text2Live~\cite{bar2022text2live} divides videos into layers and enables separate editing of each layer based on textual descriptions.

\paragraph{Parameter-Efficient Transfer Learning}
In the field of NLP, parameter-efficient fine-tuning techniques~\cite{houlsby2019parameter, hu2021lora, lester2021power, li2021prefix, he2021towards,zaken2021bitfit, sung2021training} were initially proposed to address the heavy computation of full fine-tuning large language models for various downstream tasks. These techniques aim to reduce the number of trainable parameters, thereby lowering computation costs, while still achieving or surpassing the performance of full fine-tuning.
Recently, parameter-efficient transfer learning has also been explored in the field of computer vision~\cite{jia2022visual, bahng2022exploring, chen2022adaptformer, jie2022convolutional, gao2022visual, yang2023aim, st-adapter, multimodaladapter,ssl3d, mpcn}. These methods mainly focus on adapting models within simple classification or detection tasks. In contrast, our approach focuses on adapting an T2I model for T2V generation task.

\paragraph{Temporal Shift Module}
TSM~\cite{tsm} pioneered the introduction of the temporal shift module for action recognition, employing a partial channel shift along the temporal dimension. This approach seamlessly integrates temporal cues from both preceding and succeeding frames into the current frame without incurring additional computational overhead. Subsequently, TokShift~\cite{tokenshift} implemented channel shifting along the temporal dimension for transformer architectures.
TPS~\cite{tps} further shifted patches instead of channels to model the temporal correlations.
However, such direct patch shifting would lead to inconsistency in generation task.
Additionally, Latent-shift~\cite{latentshift} and TSB~\cite{shiftgan} adapted shift module as TSM~\cite{tsm} within convolution blocks for video generation tasks.
In this work, our latent-shift attention (LSA) employs the patch-level shifting manner. In contrast to TPS, we further propose to involve all tokens in current frame as the keys and values, which guarantees the temporal consistency during generation and significantly improves the video quality.

\begin{figure*}
    \centering
    \includegraphics[width=1.0\textwidth]{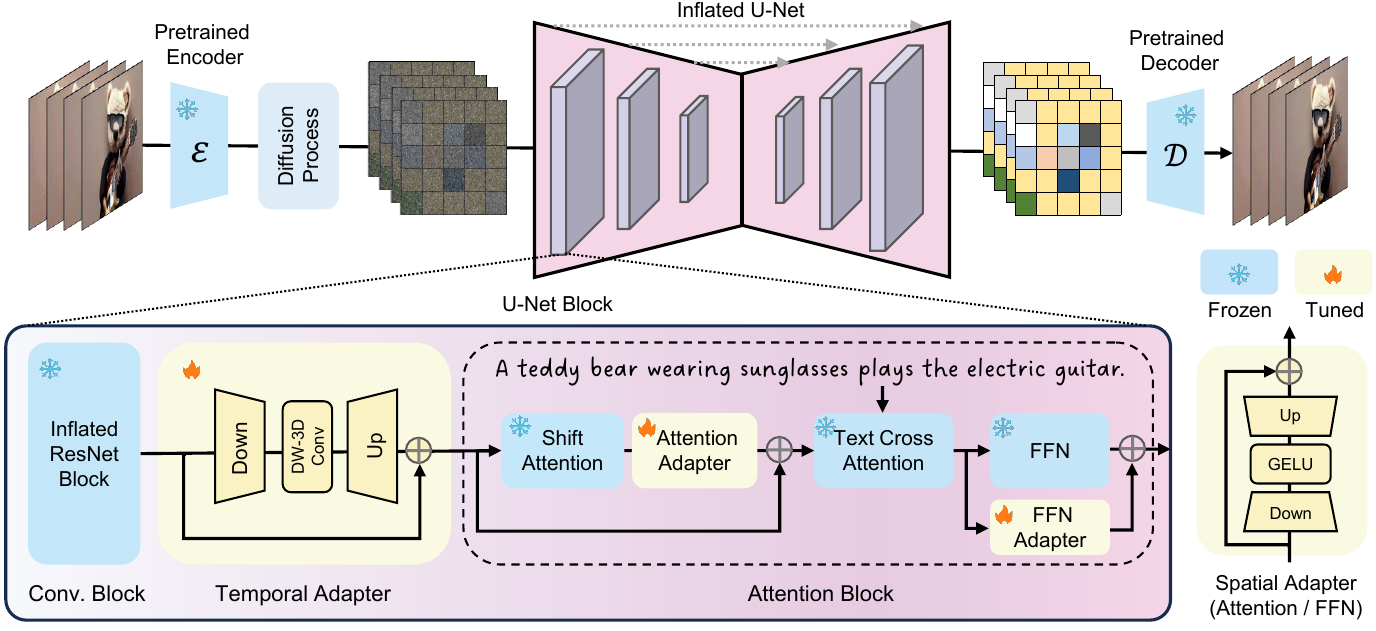}
    \caption{Pipeline of our Parameter-Efficient Text-to-Video Framework. We utilize the pre-trained auto-encoder as in Stable Diffusion~\cite{stablediffusion} to obtain latent representation. During training, we only update the parameters of newly added adapter module, highlighted in \textcolor{myred}{yellow}. The parameters of other modules are frozen, highlighted in \textcolor{mygreen}{blue}. }
    \label{fig:pipeline}
\end{figure*}

\section{Method}
In this section, we first introduce the preliminaries of Latent Diffusion Model~\cite{stablediffusion} in Sec.~\ref{preliminaries}. The pipeline of SimDA is described in Sec.~\ref{Pipeline}. Then we detail the proposed spatial and temporal adapters as well as latent-shift attention in Sec.~\ref{model}. Finally, we introduce the super resolution and text-guided video editing model in Sec.~\ref{editing}.

\subsection{Preliminaries of Stable Diffusion}
\label{preliminaries}
In this subsection, we introduce the preliminaries of Stable Diffusion~\cite{stablediffusion} model.
It is a latent diffusion model that operates in the latent space of an autoencoder $\mathcal{D}(\mathcal{E}(\cdot))$, where $\mathcal{E}$ is the encoder and $\mathcal{D}$ is the decoder.
In this model, for an image $I$ with its corresponding latent feature $\bm{x}_0=\mathcal{E}(I)$, the diffusion forward process involves iteratively adding noise to the latent space.
\begin{equation}
  q(\bm{x}_t|\bm{x}_{t-1})=\mathcal{N}(\bm{x}_t;\sqrt{\alpha_t}\bm{x}_{t-1}, (1-\alpha_t)\mathbf{I}),
\end{equation}
where $t\in\{1,...,T\}$ is the time step, $q(\bm{x}_t|\bm{x}_{t-1})$ is the conditional density of $\bm{x}_t$ given $\bm{x}_{t-1}$, $\mathbf{I}$ is identity matrix, and $\alpha_t$ is hyperparameter.
Alternatively, we can directly sample $\bm{x}_t$ at any time step from $\bm{x}_0$ with,
\begin{equation}\label{eq:forward_sample}
  q(\bm{x}_t|\bm{x}_{0})=\mathcal{N}(\bm{x}_t;\sqrt{\bar{\alpha}_t}\bm{x}_0, (1-\bar{\alpha}_t)\mathbf{I}),
\end{equation}
where $\bar{\alpha}_t=\prod_{i=1}^t\alpha_i$.

In the diffusion backward process, a U-Net denoted as $\bm{\epsilon}_\theta$ is trained to predict the noise in the latent space, aiming to iteratively recover $\bm{x}_0$ from $\bm{x}_T$. In this process, as the diffusion progresses and approaches a large value of $T$, $\bm{x}_0$ is completely disrupted and the latent representation $\bm{x}_T$ approximates a standard Gaussian distribution. Consequently, the U-Net $\bm{\epsilon}_\theta$ is trained to infer meaningful and valid $\bm{x}_0$ from random Gaussian noises. The training object can be simplified as,
\begin{equation}
\mathbb{E}_{\bm{x},\bm{\epsilon} \sim \mathcal{N}(\bm{0},\mathbf{I}),t}[\lvert\lvert \bm{\epsilon} - \bm{\epsilon}_\theta(\bm{x}_t,\bm{c},t)\rvert\rvert^2_2],
\end{equation}
where $\bm{c}$ is the embedding of condition text.

During the inference stage, it samples a valid latent representation $\bm{x}_0$ from the standard Gaussian noise $\bm{x}_T=\bm{z}_T, \bm{z}_T\sim\mathcal{N}(\bm{0}, \mathbf{I})$ using DDIM~\cite{ddim} sampling. Then, the model can decode $\bm{x}_0$ using the decoder $\mathcal{D}$ to generate the final image $I=\mathcal{D}(\bm{x}_0)$. This process could generate diverse and high-quality images based on the sampled latent representations. In contrast, our method focus on more challenge high-quality video generation.

\begin{table*}[]
\centering
    \caption{Model size and inference speed comparisons. The speed is measured in seconds on one A100 (80GB) GPU. The majority of results are sourced from~\cite{latentshift}.}
    \scalebox{0.77}{
    \begin{tabular}{l|c|c|c|c|c|c|c|c|c}
    \toprule
    \multirow{2}*{\bf Method} & \multicolumn{8}{c|}{\bf Parameters (Billion)} & \multirow{2}*{\bf Speed (s)} \\ \cline{2-9}
    & T2V Core & Auto Encoder & Text Encoder & Prior Model & Super Resolution & Frame Interpolation & Overall &Tuned & \\
    \midrule
    CogVideo~\cite{hong2022cogvideo} & $7.7$ & $0.10$ & $-$ & $-$ & $-$ & $7.7$ & $15.5$ &$15.5 $ &  $434.53$  \\ 
    Make-A-Video~\cite{singer2022make} & $3.1$ & $-$ & $0.12$ & $1.3$ & $1.4+0.7$ & $3.1$ & $9.72$ & $9.72 $ & $-$  \\
    Imagen Video~\cite{ho2022imagenvideo} & $5.6$ & $-$ & $4.6$ & $-$ & $1.2+1.4+0.34$ & $1.7+0.78+0.63$ & $16.25$ &$16.25 $ & $-$  \\
    Video LDM~\cite{videoLDM}  & $1.51$  & $0.08$  &$0.12$ &$-$  &$0.98$  &$1.51$  &$4.20$  &$2.65$  &$-$ \\
    Latent-VDM~\cite{latentshift} & $0.92$ & $0.08$ & $0.58$ & $-$ & $-$ & $-$ & ${1.58}$ & ${0.92 }$  & $ {28.62}$  \\
    Latent-Shift~\cite{latentshift} & $0.88$ & $0.08$ & $0.58$ & $-$ & $-$ & $-$ & $ 1.53$ & $ 0.88 $ & $ 23.40$  \\
    LVDM~\cite{lvdm} & $0.96$ & $0.08$ & $0.12$ & $-$ & $-$ & $-$ & $ 1.16$ & $ 1.04 $ & $ 21.23$  \\ \hline
    SimDA (Ours) & $0.88$ & $0.08$ & $0.12$ & $-$ & $-$ & $-$ & $\bf1.08$ & $\bf 0.025 $ &\bf $11.20$   \\
    \bottomrule
    \end{tabular}
    }
    \vspace{-1mm}
    \label{tab:model_size}
\end{table*}

\subsection{Pipeline}
\label{Pipeline}
Our SimDA, as shown in Fig.~\ref{fig:pipeline}, is built upon the previously introduced Stable Diffusion~\cite{stablediffusion}.
For a video clip with $t$ frames, denoted as $\{I_i\}_{i=1}^t$, we first pass it through a pre-trained encoder $\mathcal{E}$ to obtain the corresponding latent feature $\{\bm{x}_i\}_{i=1}^t$. We then input the latent features to the forward diffusion process, where noise is incrementally added to the latents. In the backward diffusion process, we utilize an inflated U-Net architecture to predict the noise for the noisy video latents. 
Specifically, for the Convolution block, we inflate the 2D ResNet~\cite{resnet} block to a 3D block to accommodate video inputs. Additionally, we incorporate a lightweight Temporal Adapter module for temporal modeling. In the Attention block, we employ a latent-shift attention mechanism for spatial self-attention and introduce two spatial adapter modules to facilitate the transfer of video information. Further details will be presented in Sec.~\ref{model}. During the inference stage, we employ DDIM~\cite{ddim} sampling to progressively denoise the latent representation sampled from a standard Gaussian distribution. Finally, we utilize a pre-trained decoder $\mathcal{D}$ to reconstruct the video from the denoised latents.

\subsection{Modeling}
\label{model}

In this section, we describe the proposed Spatial Adapter, Temporal Adapter, and Latent-Shift Attention in detail, which are the key components of our model.
\paragraph{Spatial Adapter}
The large-scale text-image pre-trained T2I model exhibits significant transferability, as evidenced by its remarkable accomplishments in tasks such as personalized T2I generation~\cite{ruiz2023dreambooth,mou2023t2iadapter} and image editing~\cite{hertz2022prompt2prompt, controlnet}. Consequently, we believe that employing a lightweight fine-tuning approach can effectively harness spatial information in the realm of video generation.
Inspired by efficient fine-tuning techniques in NLP~\cite{hu2021lora, li2021prefix } and vision tasks~\cite{chen2022adaptformer, yang2023aim}, we adopt adapters due to their simplicity. 

In our T2V framework, we design two types of spatial adapters (\emph{i.e.}, Attention Adapter and FFN Adapter) for transferring video spatial information. As shown in the bottom right of Fig.~\ref{fig:pipeline}, both adapters employ a bottleneck architecture consisting of two fully connected (FC) layers with an intermediate activation layer. The first FC layer maps the input to a lower-dimensional space, while the second FC layer maps it back to the original dimensional. Formally, for an input feature matrix $\mathbf{X}\in \mathbb{R}^{N\times d} $, the spatial adapter could be written as:
\begin{equation}
    \texttt{S\text{-}Adapter}(\mathbf{X}) = \mathbf{X} + \mathbf{W}_{\texttt{up}}(\texttt{GELU}(\mathbf{W}_{\texttt{down}}(\mathbf{X}))),
\end{equation}
where $\mathbf{W}_{\texttt{up}}$ and $\mathbf{W}_{\texttt{down}}$ are the learnable matrix with dimension $d \times l$ and $l \times d$, $l<d$.
To preserve the structure of the original network and the pretrained weights, we initialize the second FC layer $\mathbf{W}_{\texttt{down}}$ with zeros. To adapt to the spatial features of videos, we incorporate the adapter after the latent-shift attention layer. Additionally, we observe that adding adapter to the feed forward network (FFN) also helps the network transfer spatial information to videos. We will provide examples in Sec.~\ref{sec:ablation}. During training, all layers of the attention block are fixed, and only the adapters are updated.

\paragraph{Temporal Adapter}
While the spatial adapter effectively transfers spatial information to video data, modeling temporal information is crucial for T2V generation tasks. Previous approaches incorporate temporal convolution or temporal attention~\cite{videoLDM,videofactory, singer2022make} modules to capture temporal relationships. Although these modules are effective in modeling temporal dynamics, they often come with a huge number of parameters and high-dimensional input feature, resulting in significant computational and training costs.

To address this issue, we utilize the temporal adapter module for temporal modeling as~\cite{multimodaladapter, st-adapter}. In contrast to conventional spatial adapter modules, the temporal adapter module employs depth-wise 3D convolution instead of an intermediate activation layer~\cite{GELU}. The temporal adapter could be formally written as:
\begin{equation}
    \texttt{T\text{-}Adapter}(\mathbf{X}) = \mathbf{X} + \mathbf{W}_{\texttt{up}} (\texttt{3D-Conv}(\mathbf{W}_{\texttt{down}}(\mathbf{X}))).
\end{equation}

By utilizing 3D convolutions in lower-dimensional input, our approach significantly alleviates the complexity of temporal modeling. As a result, our method achieves efficient memory usage during training and exhibits the fastest inference speed among competitive approaches, as demonstrated in Table~\ref{tab:model_size}.

\paragraph{Temporal Latent-Shift Attention}

\begin{figure}
    \centering
    \includegraphics[width=1.0\linewidth]{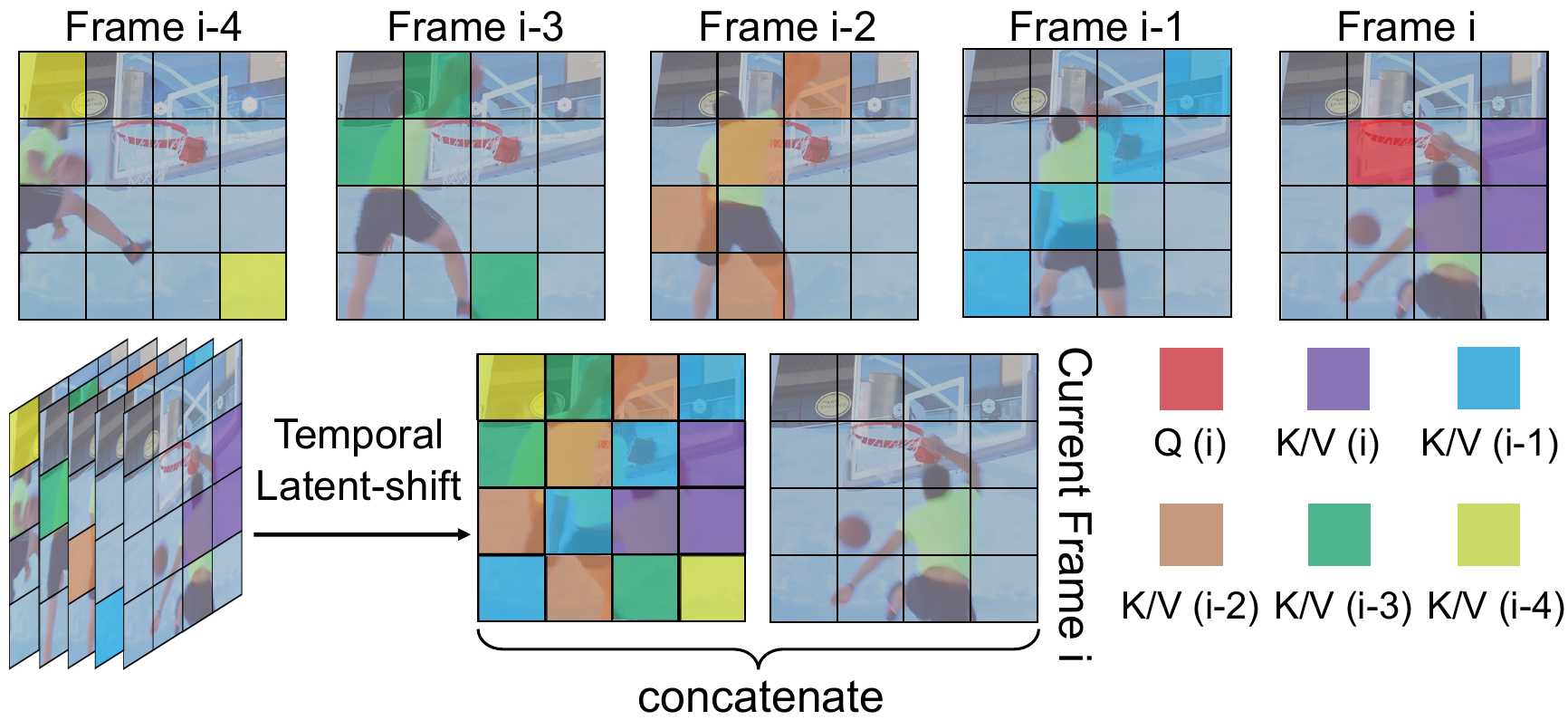}
    \vspace{-0.3cm}
    \caption{The overview of Temporal Latent-shift Attention module. It is noted that the Latent-shift attention is performed on latent space, but the visualization overview is shown on image-level for clear presentation.}
    \label{fig:shift}
    \vspace{-0.3cm}
\end{figure}

In the original T2I framework, the attention block of the U-Net only performs self-attention for individual frames, neglecting the information from other frames. While joint-space-time attention, as demonstrated in ~\cite{vivit,timesformer,xing2023svformer}, can effectively model temporal dependencies, it introduces a quadratic complexity in terms of attention calculation. 
For a video with $L$ frames and $N$ tokens, the complexity of global spatial-temporal attention becomes $O(L^2N^2)$. 
To address this issue, we propose a latent-shift attention module as shown in Fig.~\ref{fig:shift}. 
In addition to considering tokens within the current frame, we further conduct a patch-level shifting operation along the temporal dimension to shift tokens from the preceding $T$ frames onto the current frame, thereby composing a new latent feature frame.
We concatenate the latent feature of the current frame with the temporally shifted latent feature, forming the keys and values for attention calculation. The latent-shift attention can be formally written as:
\begin{equation}
    \mathbf{Q}=\mathbf{W}_\texttt{q}(\bm{x}_{z_i}), 
\end{equation}
\vspace{-0.5cm}
\begin{equation}
    \mathbf{K}=\mathbf{W}_\texttt{k}[\bm{x}_{z_i}, \bm{x}_{z_{shift}}],
\end{equation}
\vspace{-0.5cm}
\begin{equation}
    \mathbf{V}=\mathbf{W}_\texttt{v}[\bm{x}_{z_i}, \bm{x}_{z_{shift}}],
\end{equation}
where $\bm{x}_{z_i}$ denotes the query frame and $[ \cdot ]$ means concatenate.
This approach reduces the complexity of attention to $O(2LN^2)$, significantly lowering the computational burden compared to global attention. Moreover, it allows the model to learn the relationships between adjacent frames, ensuring better temporal consistency in video generation.

\begin{figure}
    \centering
    \includegraphics[width=1.0\linewidth]{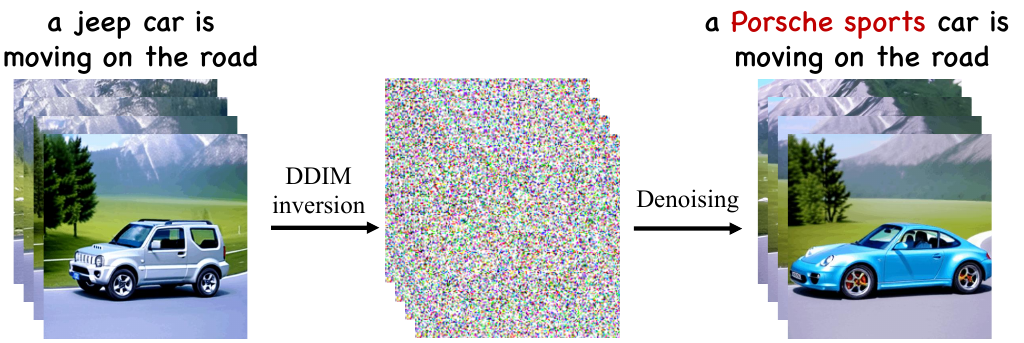}
    \caption{During inference, we sample a novel video from the latent noise inverted from the input video, guided by an edited prompt (\emph{e.g.}, ``a Porsche sports car is moving on the road"). }
    \label{fig:inference}
\end{figure}

\begin{figure*}
    \centering
    \includegraphics[width=1.0\textwidth]{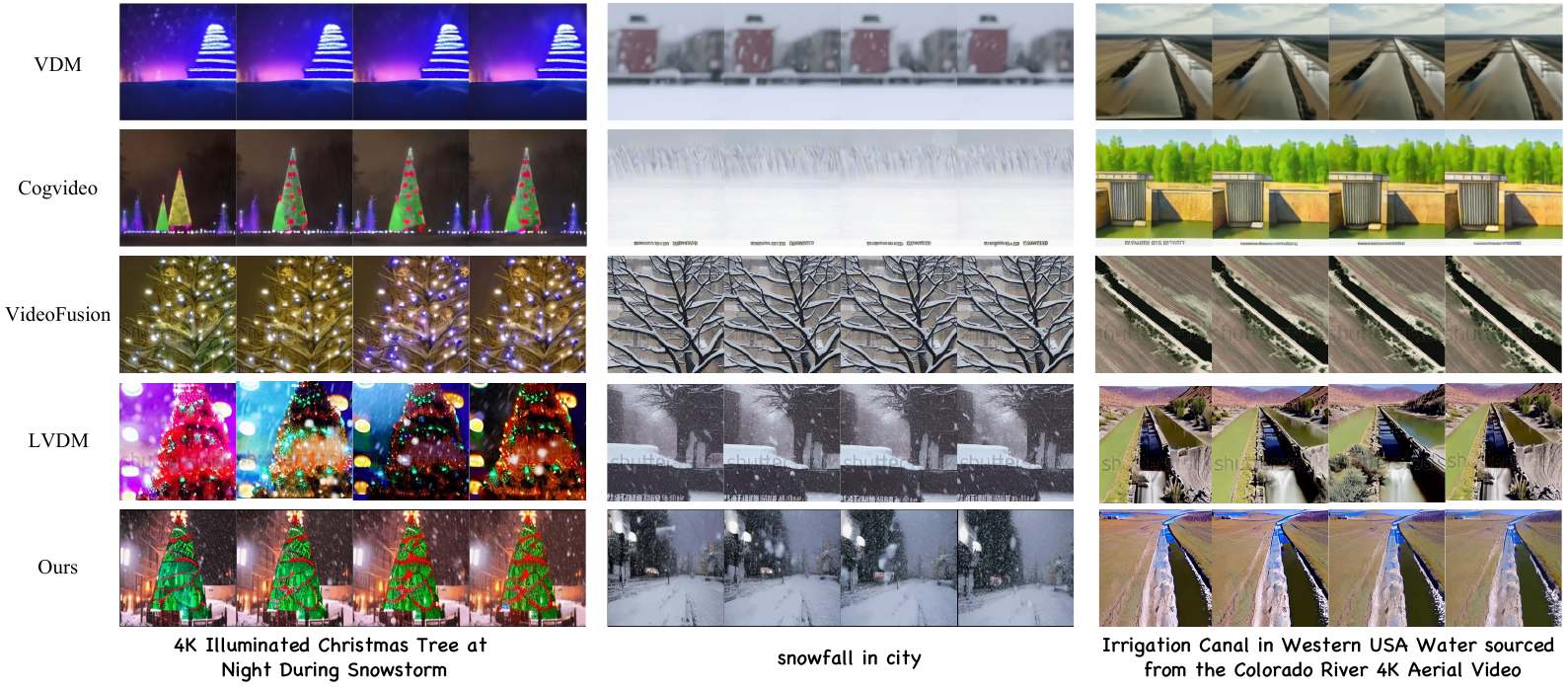}
    \vspace{-0.3cm}
    \caption{Text-to-Video generation comparison with VDM~\cite{vdm} CogVideo~\cite{hong2022cogvideo}, VideoFusion~\cite{videofusion} and LVDM~\cite{lvdm} on the user study evaluation set.}
    \label{fig:contra}
    \vspace{-0.3cm}
\end{figure*}

\subsection{Super Resolution and Editing Models}
\label{editing}

\begin{table*}[]
\centering
\caption{Text-to-Video generation comparison on MSR-VTT~\cite{msrvtt} dataset. We report the Fréchet Video Distance (FVD) scores and CLIPSIM scores.}
\vspace{-0.3cm}
\scalebox{0.9}{
\begin{tabular}{@{}ccccccc@{}}
\toprule
\bf Method       & \bf Training Data           & \bf Resolution &\bf Zero-shot &\bf Params(B) & \bf FVD($\downarrow$)   &  \bf CLIPSIM($\uparrow$)   \\ \midrule
GODIVA~\cite{wu2021godiva}       & MSR-VTT        & 128x128    & No  & -  &-  & 0.2402  \\
NÜWA~\cite{wu2022nuwa}         & MSR-VTT         & 128x128    & No & 0.87 &-  & 0.2439  \\ \midrule

\color{dt}Make-A-Video~\cite{singer2022make} & WebVid-10M + HD-VILA-10M         & 256x256    & Yes  &9.72 & -  &\color{dt} 0.3049  \\
\color{dt}VideoFactory~\cite{videofactory}  & WebVid-10M + HD-VG-130M &256x256   &Yes & 2.04  &-  & \color{dt}0.3005 \\ 
LVDM~\cite{lvdm}         & WebVid-2M   & 256x256    &Yes   &1.16  &742   & 0.2381  \\
MMVG~\cite{mmvg}  & WebVid-2.5M   & 256x256    &Yes   &-  &-   & 0.2644  \\      
CogVideo~\cite{hong2022cogvideo}     & WebVid-5.4M         & 256x256    & Yes &15.5  &1294  & 0.2631  \\
ED-T2V~\cite{edt2v}    &WebVid-10M   & 256x256 &Yes & 1.30  & - & 0.2763 \\
MagicVideo~\cite{magicvideo}  &WebVid-10M & 256x256 &Yes & -  & 998 & - \\
Video-LDM~\cite{videoLDM}    & WebVid-10M         & 256x256    & Yes  &4.20  &-  & 0.2929  \\
VideoComposer~\cite{videocomposer}    & WebVid-10M     & 256x256 & Yes & 1.85  & 580  & 0.2932 \\
Latent-Shift~\cite{latentshift} & WebVid-10M & 256x256    & Yes   &1.53 &- & 0.2773  \\
VideoFusion~\cite{videofusion}   & WebVid-10M         &256x256     &Yes    &  1.83  &581   &0.2795  \\ \midrule
SimDA (Ours)         & WebVid-10M       & 256x256    & Yes  &\textbf{1.08} &\bf 456  & \textbf{0.2945}  \\ \bottomrule
\end{tabular}
}
\label{Tab:msrvtt}
\vspace{-0.1em}
\end{table*}

\paragraph{Super Resolution (SR)} Due to constraints of limited GPU memory and the lack of high-resolution video-text datasets, most existing methods~\cite{lvdm, videoLDM, latentshift}, including ours, are only able to generate images at a resolution of $256\times256$. To overcome this limitation and generate higher-resolution outputs, we adopt two-stage training approach similar to cascaded Diffusion Models~\cite{cascaded}. In the first stage, we generate videos with a  $256\times 256$ resolution using our SimDA methods. In the second stage, we employ an LDM upsampler~\cite{stablediffusion} to enhance the resolution of the videos to $1024\times1024$. We incorporate noise augmentation and noise level conditioning, and train a super-resolution model using the following equation:
\begin{equation}
\mathbb{E}_{\bm{x},\bm{\epsilon} \sim \mathcal{N}(\bm{0},\mathbf{I}),t}[\lvert\lvert \bm{\epsilon} - \bm{\epsilon}_\theta([\bm{x}_t,\bm{x}_{low}],\bm{c},t)\rvert\rvert^2_2],
\end{equation}
where $\bm{x}_{low}$ is the low-resolution video, we concatenate it with $\bm{x}_t$ frame by frame following Video LDM~\cite{videoLDM}. The architecture of SR is similar to T2V model in first stage, we change the original U-Net block by adding Spatial and Temporal Adapters as described in Sec.~\ref{model} and only finetune the new added modules.

\paragraph{Text-guided Video Editing} In addition to performing T2V generation, our method could turn into one-shot tuning for text-guided video editing following Tune-A-Video~\cite{tuneavideo}. The training pipeline of editing model is the same to our T2V method. However, for the inference stage, we adopt the DDIM inversion latents instead of random noisy latents together with edited prompt for novel video generation as shown in Fig.~\ref{fig:inference}. By doing so, the pixel-level information control could remain in the inversion latent as demonstrated in ~\cite{tuneavideo}. Owing to the light-weight module and efficient pipeline of our method, SimDA needs fewer training steps (200 steps compared to 500 steps) and thus the training time and inference time is much faster than Tune-A-Video~\cite{tuneavideo}.

\section{Experiments}

\begin{figure*}
    \centering
    \includegraphics[width=0.85\textwidth]{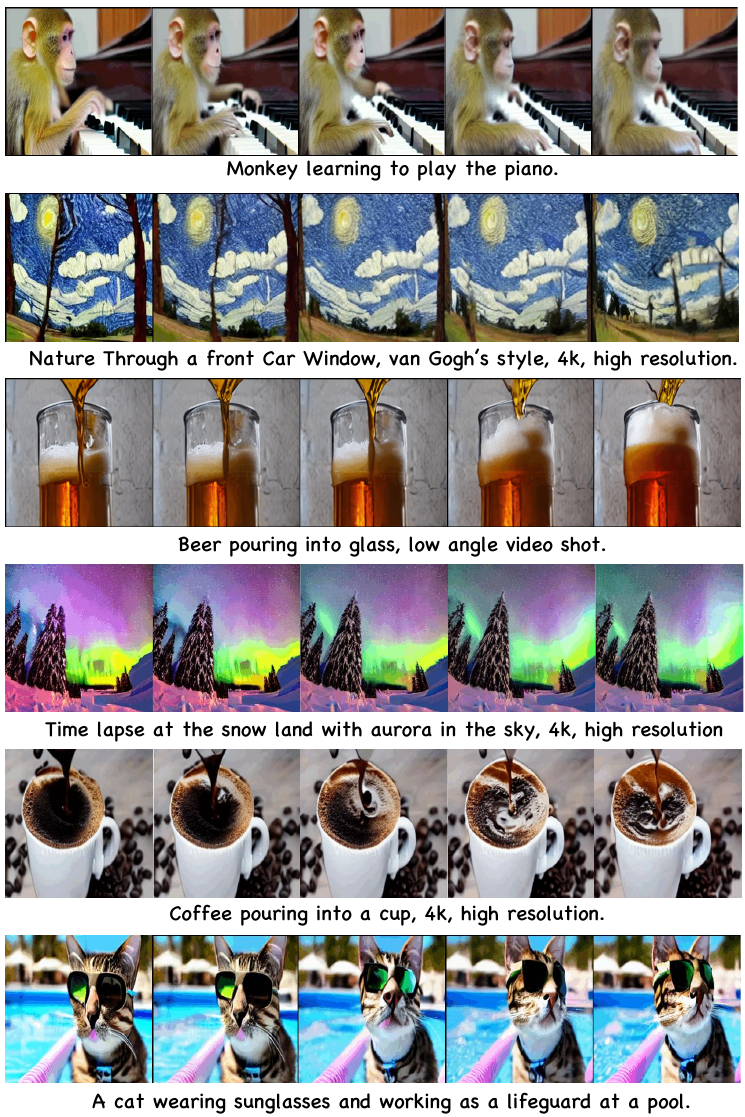}
    \caption{Results of extending our SimDA to text-to-video generation.}
    \label{fig:text2video}
\end{figure*}

\subsection{Implementation Details}
Our T2V method is composed of two-stage models. The first model predicts video frames with a resolution $256\times256$ (with a latent size of $32\times32$), while the second model is a 4$\times$ upsampler, producing a resolution of $1024\times1024$. 
We train the general T2V model on WebVid-10M~\cite{webvid} dataset following~\cite{videoLDM,latentshift}. We follow previous methods~\cite{videofactory, latentshift, magicvideo} to report the CLIP score~\cite{clip} and FVD (Fréchet Video Distance) score~\cite{fvd} on MSR-VTT~\cite{msrvtt}. Besides, we compare the FVD score and CLIP score on evaluation set of WebVid~\cite{webvid} as in VideoFactory~\cite{videofactory}. We also compare the parameter scale and inference speed of our method with some open-sourced methods~\cite{lvdm,videofusion, latentshift}. Finally, we also provide a user study between our work and VDM~\cite{vdm}, Latent-shift~\cite{latentshift}, Video-Fusion~\cite{videofusion} and LVDM~\cite{lvdm}. 

\subsection{Evaluation on Text-to-Video Generation}
To fully evaluate the generation performance of our SimDA, we conduct automatic evaluations on two distinct datasets: WebVid~\cite{webvid} (Val), which shares the same domain as the training data, and MSR-VTT ~\cite{msrvtt} in a zero-shot setting.

\paragraph{Evaluation on MSR-VTT}
As shown in Table~\ref{Tab:msrvtt}, we evaluate CLIPSIM~\cite{clip} and FVD~\cite{fvd} on the widely used video generation benchmarks, MSR-VTT~\cite{msrvtt}. We randomly select one text prompt per example from MSR-VTT~\cite{msrvtt} and generate a total of 2,990 videos. Despite being a zero-shot setting, our method achieves an average CLIPSIM of 0.2945 that surpasses most of the competitors, indicating a strong semantic alignment between the generated videos and the input text. 
Though Make-A-Video~\cite{singer2022make} and VideoFactory~\cite{videofactory} offer higher CLIP scores, they utilize additional large-scale HD-VILA~\cite{hd-vila} datasets for training.

\paragraph{Evaluation on WebVid}
As shown in Table~\ref{Tab:webvid}, we create a validation set consisting of 4,476 randomly extracted text-video pairs from WebVid-10M. These pairs are not included in the training data following~\cite{videofactory}. We conduct evaluations on this validation set and obtain impressive results. Our method achieves an FVD score of 363.98 and a CLIPSIM score of 0.3054. These scores are significantly higher than those achieved by existing methods such as VideoFusion~\cite{videofusion} and LVDM~\cite{lvdm}. Besides, our method shows competitive results compared to VideoFactory~\cite{videofactory} which is trained with much larger datasets. These results clearly demonstrate the superiority of our approach.

\paragraph{Human Evaluation}
In order to address the limitations of existing metrics and assess the performance of our SimDA from a human perspective, we conduct an extensive user study. The study involves comparing our method with four state-of-the-art methods. Specifically, we select two publicly available models, namely VideoFusion~\cite{videofusion} and LVDM~\cite{lvdm}. Additionally, we consider two methods with similar scale parameters, VDM~\cite{vdm} and Latent-shift~\cite{latentshift}, which only showcase some samples on their websites.

For each case, participants were provided with two video samples, one is generated by our method and the other is from a competitor. They were then asked to compare the two samples in terms of video quality and text-video similarity. To ensure fairness in the comparisons, we also report the ratio of network parameter compared to ours.
The results, along with the parameter ratios, are presented in Table~\ref{Tab:user}. The user study approach allows us to gain in-depth insights into the subjective evaluation of our method.

\begin{table}[]\small
\centering
\caption{Text-to-video generation on the validation set of WebVid~\cite{webvid}. We report the FVD and CLIPSIM scores.}
\vspace{-0.2cm}
\begin{tabular}{@{}lccc@{}}
\toprule
\bf Method    &\bf Params(B)   & \bf FVD($\downarrow$)    & \bf CLIPSIM($\uparrow$) \\ \midrule
LVDM~\cite{lvdm}      &1.16   & 455.53 & 0.2751  \\
VideoFusion~\cite{videofusion} &1.83  & 414.11 & 0.3000  \\
\color{dt}VideoFactory~\cite{videofactory} &2.04   & \color{dt}292.35 & \color{dt}0.3070  \\ \midrule
SimDA (Ours)      &\bf 1.08   & \bf 363.98       & \bf 0.3054        \\ \bottomrule
\label{Tab:webvid}
\vspace{-0.3cm}
\end{tabular}
\end{table}

\paragraph{Qualitative Results} The visualization of T2V generation results are shown in Fig.~\ref{fig:text2video} and  Fig.~\ref{fig:fig1}(a). Besides, we show the comparison results in  Fig.~\ref{fig:contra}.
More examples can be found at our website.

\begin{table}[]
\centering
\caption{User preference is depicted as a percentage indicating the proportion of individuals favoring our method over the compared approach. Param Ratio means the ratio of the network parameter \emph{v.s.} Ours.}
\scalebox{0.8}{
\begin{tabular}{@{}ccccc@{}}
\toprule
\multicolumn{1}{l}{\bf Sample}                                                   &\bf Method       &\bf Param Ratio &\bf Quality &\bf Faithfulness \\ \midrule
\multirow{2}{*}{\begin{tabular}[c]{@{}c@{}}Open\\  Website\end{tabular}}     & VDM~\cite{vdm}          & $0.83\times$            & 85.2\%              &  81.4\%                    \\ \cmidrule(l){2-5} 
                                                                             & Latent-Shift~\cite{latentshift} & $1.41\times$          &  81.5\%             &    79.3\%                   \\ \midrule
\multirow{2}{*}{\begin{tabular}[c]{@{}c@{}}Pretrained\\  Model\end{tabular}} & VideoFusion~\cite{videofusion}  &  $1.69\times$        & 78.3\%              & 79.5\%                     \\ \cmidrule(l){2-5} 
                                                                             & LVDM~\cite{lvdm}         & $1.07\times$            & 83.4\%              & 84.7\%                     \\ \bottomrule
\label{Tab:user}
\end{tabular}
}
\end{table}

\begin{table*}[t]
\caption{Quantitative comparison with evaluated baseline~\cite{tuneavideo}. The ``Training'' refers to the process of optimization, and ``Memory'' refers to the GPU memory.}
\vspace{-2mm}
\centering
\setlength\tabcolsep{0.3pt}
\small
\scalebox{1.06}{
\begin{tabular}{c|cc|cc|cc|cc|c}
\toprule
\multirow{2}{*}{\textbf{Method}} & \multicolumn{2}{c|}{\textbf{Frame consistency}} &  \multicolumn{2}{c|}{\textbf{Textual alignment}} & \multicolumn{2}{c|}{\textbf{Runtime} [min]} & \multicolumn{2}{c|}{\textbf{Memory} [Gib]} & \textbf{Params} [Mb]  \\ \cline{2-10}
              & ~{\footnotesize CLIP Score$\uparrow$} ~~& {\footnotesize User Vote$\uparrow$}~ & ~{\footnotesize CLIP Score$\uparrow$}~~ & {\footnotesize User Vote$\uparrow$}~ & ~{\footnotesize Training$\downarrow$}~~ & {\footnotesize Inference$\downarrow$}~  & ~{\footnotesize Training$\downarrow$} ~~& {\footnotesize Inference$\downarrow$}~ &
              {\footnotesize Tuned$\downarrow$} \\ \hline
Tune-A-Video~\cite{tuneavideo}~      & 94.1      &  31.2\%           &31.8      &  39.5\%  & 9.3 & 0.8 & 31.3 & 11.4  &74.4     \\ 

SimDA(Ours)            & \bf 94.9 & \bf 68.8\% & \bf 31.9  & \bf 60.5\%  & \bf 2.5 & \bf 0.4 &\bf 28.6 &\bf 8.8 &\bf 24.9 \\ \bottomrule
\end{tabular}%
}
\label{tab:tuneavideo}
\vspace{-2mm}
\end{table*}

\paragraph{Parameter Size and Inference Speed}
We conduct a comparison of number of parameters and inference speed, and the results are presented in Table~\ref{tab:model_size}. For the speed comparison, we select CogVideo~\cite{hong2022cogvideo}, Latent-Shift~\cite{latentshift} and LVDM~\cite{lvdm}.
SimDA, on the other hand, stands out as it is significantly smaller than previous works and exhibits faster inference speed compared to other methods. Despite having fewer parameters, SimDA achieves superior performance in various benchmarks when compared to other methods. This validation further highlights our advantages in terms of model efficiency and performance.

\subsection{Evaluation on Text-guided Video Editing}
Following the methodology of previous studies~\cite{tuneavideo}, we employ CLIP score~\cite{clip} and a user study to evaluate the performance of different methods in terms of frame consistency and textual alignment.

First, we calculate the CLIP image embedding for all frames in the edited videos to measure frame consistency. The average cosine similarity between pairs of video frames is reported. Additionally, to assess textual faithfulness, we compute the average CLIP score between frames of the output videos and the corresponding edited prompts. A total of 15 videos from the dataset~\cite{davis} were selected and edited based on object, background and style, resulting in 75 edited videos for each model. 
The average results, presented in Table~\ref{tab:tuneavideo}, highlight our method's exceptional ability to achieve semantic alignment. 

Secondly, we conduct a user study involving videos and text prompts. Participants were asked to vote for the edited videos that exhibited the best temporal consistency and those most accurately matched the textual description. Table~\ref{tab:tuneavideo} demonstrates that our method, SimDA, receives the highest number of votes in both aspects, indicating superior editing quality and a strong preference from users in practical scenarios.

\begin{figure}
    \centering
    \includegraphics[width=1.0\linewidth]{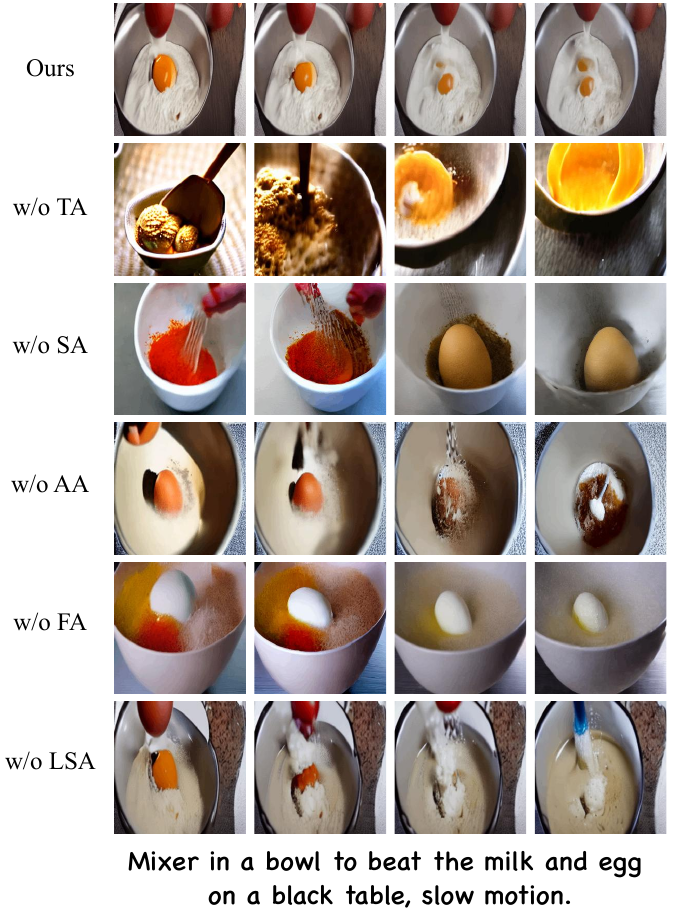}
    \caption{Ablation of T2V generation results. TA, SA, AA, FA and LSA refer to Temporal Adapter, Spatial Adapter, Attention Adapter, FFN Adapter and Latent-Shift Attention. }
    \label{fig:ablation}
\end{figure}

\subsection{Ablation Study}
\label{sec:ablation}
In this section, we will discuss the effect of our proposed modules, we perform experiments on 1$K$ samples from validation set of WebVid~\cite{webvid}.

\begin{table}[]
\centering
\caption{Ablation study on different modules. We report the FVD~\cite{fvd} and CLIPSIM~\cite{clip} on 1$K$ samples from the validation set of WebVid-10M~\cite{webvid}. TA, SA, AA, FA and LSA represent Temporal Adapter, Spatial Adapter, Attention Adapter, FFN Adapter and Latent-shift Attention, respectively.}
\scalebox{0.93}{
\begin{tabular}{@{}ccccccc@{}}
\toprule
       & \bf TA &\bf AA &\bf FA &\bf LSA & \bf FVD($\downarrow$) &\bf CLIPSIM($\uparrow$) \\ \midrule
w/o TA & & \checkmark & \checkmark & \checkmark & 1470.1    & 0.2629  \\
w/o SA & \checkmark &  &  & \checkmark & 811.3    & 0.2822    \\
w/o AA &  \checkmark &   & \checkmark & \checkmark &  764.8     &  0.2851        \\
w/o FA & \checkmark & \checkmark &    & \checkmark   &623.7     &  0.2962        \\
w/o LSA &  \checkmark & \checkmark & \checkmark &                 &618.2     & 0.3011       \\ \midrule
Ours   &   \checkmark  &   \checkmark                 &    \checkmark          &   \checkmark       &\bf 530.2     & \bf 0.3034        \\ \bottomrule
\end{tabular}
}
\label{Tab:ablation}
\end{table}

\paragraph{Effect of Temporal Adapter}
Temporal modeling is a crucial component of video generation. In our video editing task, when compared to methods that rely on temporal attention modeling like Tune-A-Video~\cite{tuneavideo}, we observe that our temporal adapter is more lightweight and achieves superior editing results as in Table~\ref{tab:tuneavideo}. Additionally, we conduct ablation experiments, as shown in Table~\ref{Tab:ablation} and Fig.~\ref{fig:ablation}, where the lack of Temporal Adapter (TA) results in significantly higher FVD score and chaotic temporal sequences in the generated videos.

\paragraph{Effect of Spatial Adapter}
We also validate the effectiveness of the Spatial Adapter (SA) in transferring spatial knowledge of videos. As shown in Table~\ref{Tab:ablation}, without the Attention Adapter (AA) and FFN Adapter (FA), the model's FVD and CLIPSIM scores for generated videos will become worse. Additionally, it can be observed from the Fig.~\ref{fig:ablation} that the model exhibits misconceptions in understanding the text prompt without the spatial adapter.

\paragraph{Effect of Latent Shift Attention}
To investigate the impact of Latent-shift Attention (LSA) on the model, we replace it with regular single-frame spatial attention. Besides observing a decline in FVD and text alignment CLIPSIM scores in Table~\ref{Tab:ablation}, we also test the CLIPSIM of each frame within the same video, which decreased from 96.4 to 94.5. This demonstrates that our LSA module can effectively model the adjacent frames relationship, leading to more consistent videos.

\section{Conclusion}
In this paper, we proposed SimDA, a parameter efficient video diffusion model for text guided video generation and editing. With the proposed light-weight spatial and temporal adapters, our method not only transferred from powerful spatial information but also modeled temporal relationship with least new parameters. The experimental results demonstrated that our approach has the fastest training and inference speed, while maintaining competitive generation and editing results. Our work is the first parameter-efficient video diffusion method serving as an efficient T2V fine-tuning baseline and paved the way for future research.

{\small
\bibliographystyle{ieee_fullname}
\bibliography{egbib}
}

\end{document}